\documentclass[10pt,twocolumn,letterpaper]{article}

\usepackage{cvpr}              
\usepackage{colortbl}
\usepackage{multirow}
\usepackage{makecell}
\usepackage{arydshln}
\definecolor{cvprblue}{rgb}{0.21,0.49,0.74}
\definecolor{paris_purple}{RGB}{81,81,171}
\definecolor{paris_red}{RGB}{218,139,186}
\definecolor{paris_orange}{RGB}{238,161,80}
\definecolor{tablecolor}{gray}{.93}
\definecolor{dm_green}{RGB}{84,130,53}
\definecolor{dm_purple}{RGB}{112,48,160}

\usepackage[pagebackref,breaklinks,colorlinks,allcolors=cvprblue]{hyperref}
\usepackage[misc]{ifsym}

\newcommand{\implus}[1]{\textcolor{black}{#1}}
\newcommand{\rewrite}[1]{\textcolor{black}{#1}}

\newcommand{\append}[1]{\textcolor{black}{#1}}

\def\paperID{6401} 
\def\confName{CVPR}
\def\confYear{2025}

\title{RSAR: Restricted State Angle Resolver and Rotated SAR Benchmark}

\author{Xin Zhang$^1$, \
        Xue Yang$^2$, \
        Yuxuan Li$^1$, \
        Jian Yang$^1$, \
        Ming-Ming Cheng$^{1,3}$, \
        Xiang Li$^{1,3\ \textrm{\Letter}}$\\
        $^1$VCIP, CS, Nankai University \ \
        $^2$Shanghai AI Laboratory \ \
        $^3$NKIARI, Shenzhen Futian\\
        {\tt\small zhasion@mail.nankai.edu.cn, yangxue@pjlab.org.cn, yuxuan.li.17@ucl.ac.uk}\\
        {\tt\small \{csjyang,cmm,xiang.li.implus\}@nankai.edu.cn}\\
}
\begin{document}
\maketitle
\begin{abstract}

Rotated object detection has made significant progress in the optical 
\implus{remote sensing.}
However, advancements in the Synthetic Aperture Radar (SAR) field are laggard behind, primarily due to the absence of a large-scale dataset. Annotating such a dataset is inefficient and costly.
A promising solution is to employ a weakly supervised model \implus{(e.g., trained with available horizontal boxes only)} to generate pseudo-rotated boxes \implus{for reference} before manual calibration.
Unfortunately, the existing weakly supervised models exhibit limited accuracy in predicting the object's angle.
Previous works attempt to enhance angle prediction by using angle resolvers that decouple angles into cosine and sine encodings.
In this work, we \implus{first} reevaluate these resolvers from a unified perspective of dimension mapping \implus{and expose that they share the same shortcomings:}
these methods overlook the unit cycle constraint inherent in these encodings, easily leading to prediction biases.
To address this issue, we propose the \textbf{Unit Cycle Resolver} (UCR), which incorporates a unit circle constraint loss to improve angle prediction accuracy.
Our approach can effectively improve the performance of existing state-of-the-art \implus{weakly supervised} methods and even surpasses fully supervised models on existing optical benchmarks (i.e., DOTA-v1.0 dataset). 
With the aid of UCR, we further annotate and introduce \textbf{RSAR}, the largest multi-class rotated SAR object detection dataset to date. 
Extensive experiments on both RSAR and optical datasets demonstrate that our UCR enhances angle prediction accuracy. Our dataset and code can be found at:~\url{https://github.com/zhasion/RSAR}.
\end{abstract}

\section{Introduction}
\label{sec:intro}
Rotated object detection~\cite{zhao2019object}, which provides more precise localization than horizontal object detection, is widely utilized in remote sensing~\cite{xie2021oriented,ding2019learning,li2023large,xia2018dota}, 3D detection~\cite{zheng2020rotation,liu2021autoshape}, and scene text detection~\cite{liu2018fots,ma2018arbitrary,zhou2017east, liao2018rotation}. Particularly in the field of remote sensing, the increased accessibility of optical satellite imagery~\cite{xia2018dota, sun2022fair1m} facilitates numerous contributions to rotated object detection. Synthetic Aperture Radar (SAR)~\cite{sun2021spaceborne}, a prominent remote sensing technology with all-weather imaging capabilities, attracts considerable attention in recent years. 
With the introduction of SAR datasets~\cite{zhang2021sar,wei2020hrsid,wang2023category}, a growing number of studies~\cite{li2024unleashing,zhou2022pvt,dai2024denodet,li2024sm3det} are focusing on SAR object detection. 

\begin{figure}[t]
    \centering
    \includegraphics[width=1.0\linewidth]{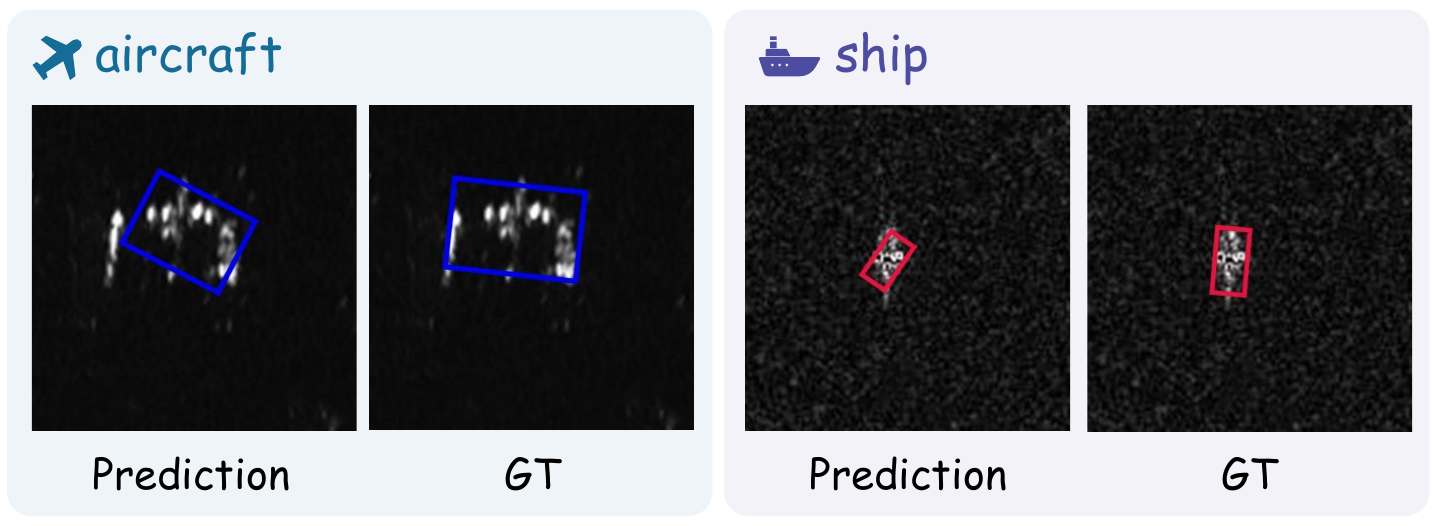}
    \caption{
    A comparison between prediction results from a weakly supervised model and the ground truth (GT). The weakly supervised model's accuracy in predicting the object angle requires further improvement.
    }
    \label{fig:compare_sar_optical}
\end{figure}

\begin{figure*}[t]
    \centering
    \includegraphics[width=1.0\linewidth]{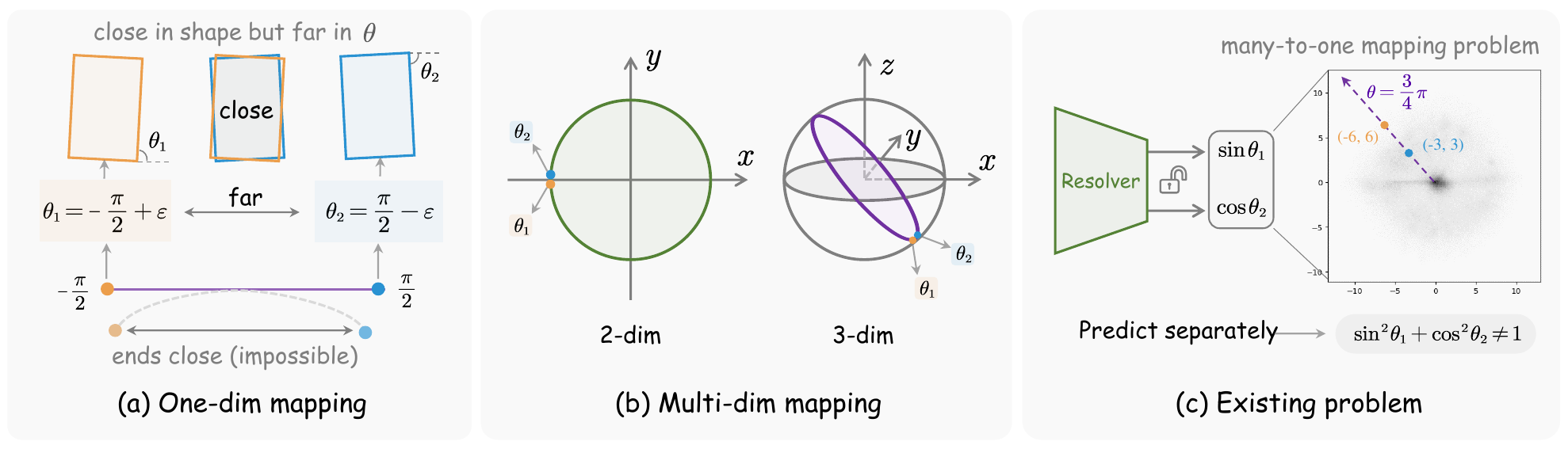}
    \caption{Diagram analysis of the angle boundary discontinuity problem from the unified perspective of \emph{dimensional mapping}. (a) One-dimensional space leads to the problem of angle boundary discontinuity. (b) Mapping one-dimensional values to two-dimensional and three-dimensional spaces helps address the issue. (c) Existing methods overlook the unit circle constraint inherent in the angle encoding states, leading to a many-to-one problem that introduces biases in model optimization and predictions. Our unified perspective clarifies the angle boundary discontinuity issue and exposes the potential shortcomings of existing methods.}
    \label{fig:angle_resolver}
    \vspace{-10pt}
\end{figure*}


Current research on SAR image interpretation primarily focuses on horizontal object detection, where performance tends to saturation. In contrast, progress in rotated SAR object detection is much slower. This is mainly due to the lack of a large-scale dataset specifically designed for rotated SAR object detection. Annotating such a dataset is both costly and inefficient, and a potential solution is to leverage a weakly supervised model.
This weakly supervised model is trained with horizontal bounding boxes and generates pseudo-rotated detection boxes, which are then manually refined for \implus{efficient} calibration.
However, as shown in Fig.~\ref{fig:compare_sar_optical}, 
accurately predicting the detected object's angle remains challenging for existing weakly supervised models.
Therefore, improving angle prediction accuracy is crucial for enhancing the applicability of weakly supervised methods in SAR datasets.

The main challenge in accurate angle prediction lies in the angle boundary discontinuity problem~\cite{yang2021rethinking}. 
To address this issue radically, previous related works~\cite{xu2024rethinking,yu2023phase} utilize an angle resolver to decompose an angle value into sine and cosine components. Although their specific formulations vary, we reevaluate them from a unified and insightful perspective of \emph{dimensional mapping}. All of these methods employ \emph{dimensional mapping} to transform the discontinuous one-dimensional angle regression task into a continuous multi-dimensional encoding states regression task. 

From this unified perspective, it is evident that previous approaches overlook the constraint that the angle encoding states must adhere to the unit circle conditions. Specifically, these methods predict multiple encoding states independently, which can lead to deviations from the unit circle constraint. This deviation complicates the optimization process and easily introduces biases in angle predictions. To address this, we propose an innovative restricted state angle resolver called Unit Cycle Resolver (UCR). It incorporates a unit cycle constraint loss to ensure the encoding states conform to the unit circle constraint. 
\implus{To justify its general effectiveness, }
we evaluate our method on the large-scale rotated object detection dataset DOTA-v1.0. 
Our method can effectively improve the performance of existing state-of-the-art weakly supervised methods and even achieves performance comparable to that of fully supervised methods.

\implus{With the initial pseudo rotated box labels provided by the weakly supervised model~\cite{yu2024h2rbox} using our proposed UCR, we efficiently construct the RSAR dataset through manual calibration.}
RSAR is a comprehensive multi-class large-scale rotated SAR object detection dataset, which comprises 95,842  SAR images and 183,534 annotated instances across six typical SAR object categories. To our knowledge, RSAR is the largest rotated object detection dataset available in this field to date. Extensive experiments on RSAR and more optical datasets demonstrate that our UCR significantly enhances model performance in angle prediction. 

Our contributions can be summarized as follows:
\begin{itemize}
[leftmargin=20pt]\setlength{\itemsep}{5pt}
    \item We analyze existing angle resolver methods and their limitations from a unified perspective, proposing an innovative resolver called Unit Cycle Resolver (UCR). This resolver ensures that angle encodings adhere to inherent constraints by incorporating a unit circle constraint loss.
    \item We efficiently construct RSAR, a large-scale multi-class rotated SAR object detection dataset, with the help of the proposed UCR and existing weakly supervised model. To our best knowledge, RSAR is the largest rotated object detection dataset in the field of SAR to date.
    \item Experiments demonstrate that our method can significantly improve the performance of previous state-of-the-art approaches in weakly supervised tasks. It significantly improves the accuracy of angle prediction in both RSAR and optical datasets. Notably, the improved method achieves performance comparable to fully supervised methods in DOTA-v1.0.
\end{itemize}

\begin{figure*}[t]
    \centering
    \includegraphics[width=0.98\linewidth]{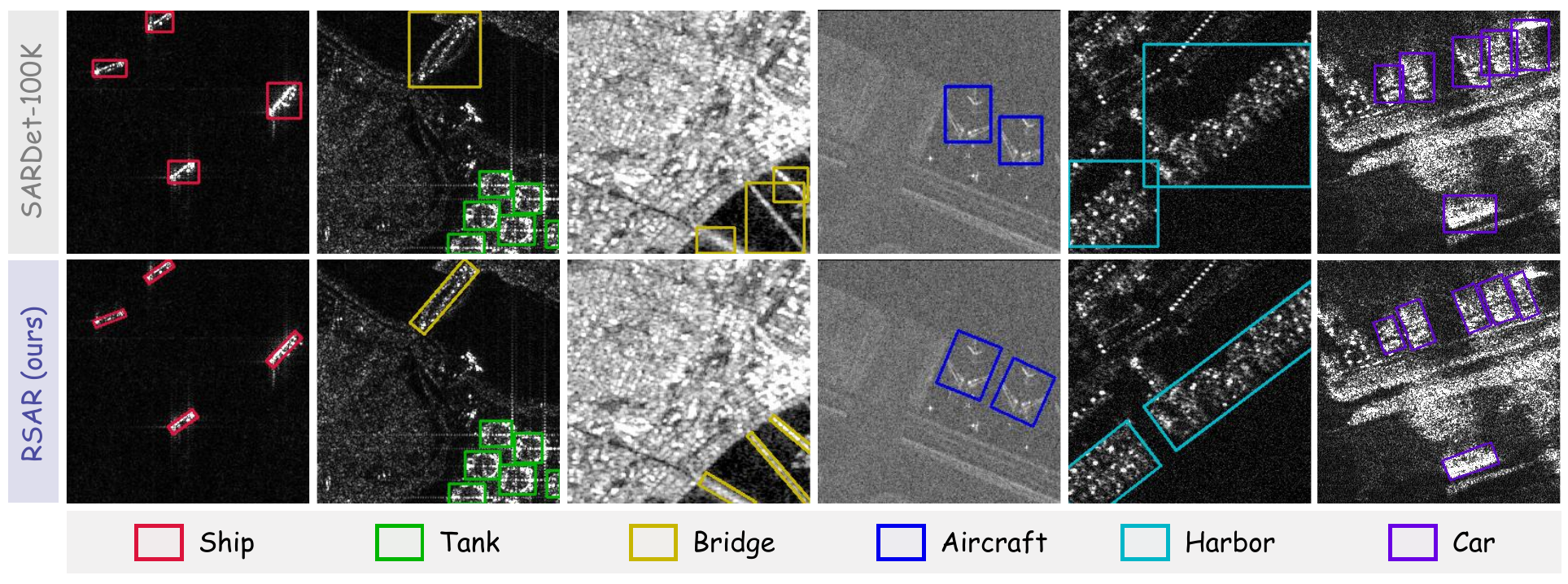}
    \caption{Visualization of images from RSAR and SARDet-100K. Rotated annotations in RSAR offer higher location accuracy compared to horizontal annotations in SARDet-100K. Rotated SAR object detection presents greater challenges than horizontal SAR object detection.}
    \label{fig:dataset_visual}
\end{figure*}

\section{Related Work}
\label{sec:related}

\textbf{SAR Object Detection:} Traditional SAR object detection methods primarily rely on manually designed features~\cite{robey1992cfar,wang2016saliency}, while deep neural networks are widely adopted in recent years~\cite{cui2019dense,zhang2020fec,chen2022geospatial}. The success of these networks largely depends on the availability of sufficient data. Early SAR object detection datasets mainly focus on ship objects. For instance, SSDD~\cite{zhang2021sar} introduces the first publicly available SAR ship detection dataset, and HRSID~\cite{wei2020hrsid} is designed specifically for ship detection in high resolution. In recent years, larger datasets and more detection categories are proposed~\cite{wang2023category,zhirui2023sar,li2024sardet}, including SARDet-100K~\cite{li2024sardet}, which provides a large-scale dataset for horizontal SAR object detection. Building on these datasets, various detection frameworks~\cite{li2024unleashing,li2017ship,cui2019dense,zhang2021quad,li2024predicting} are developed. However, current methods focus primarily on horizontal detection, where evaluation metrics reach saturation (\eg, the evaluation metric of AP$_{50}$ typically exceeding 90). In contrast, rotated object detection, which offers higher localization accuracy and presents more challenges, has progressed slowly. This delay is due to the limitations of single-category, small-scale datasets in this domain.

\noindent{\textbf{Rotated Object Detection:}}
Existing models represent a detection bounding box using the format $(cx,cy,w,h,\theta)$. The key distinction between rotated object detection and horizontal object detection lies in the prediction of angles $\theta$. Currently, most mainstream approaches build upon horizontal detection frameworks~\cite{cai2018cascade, ren2016faster, redmon2016you, li2020generalized} by incorporating angle regression predictions. 
These methods include single-stage frameworks like R$^3$Det~\cite{yang2021r3det} and S$^2$ANet~\cite{han2021align}, as well as two-stage frameworks such as ReDet~\cite{han2021redet} and SCRDet~\cite{yang2019scrdet}, along with DETR-based frameworks~\cite{zhudeformable, zeng2024ars}.
Weakly supervised tasks~\cite{yang2022h2rbox,yu2024h2rbox,luo2024pointobb,yu2024point2rbox,ren2024pointobb} also play an important part in rotated object detection, particularly when rotated bounding box labels are unavailable. H2RBox~\cite{yang2022h2rbox} addresses this challenge by utilizing horizontal boxes as weak supervision signals for detecting rotated objects. H2RBox-v2~\cite{yu2024h2rbox} further reduces the performance gap between weakly and fully supervised tasks by leveraging symmetry consistency information of the detected objects. However, the boundary discontinuity caused by angles $\theta$ persists in most rotated object detection methods.

\noindent{\textbf{Angle Boundary Discontinuity Problem:}}
Initially, using L1 loss to directly supervise the angle $\theta$ for rotated object detection results in a sharp increase in angle loss at the boundaries of the defined angle range~\cite{yang2021rethinking}. Subsequent research aims to address this issue. For instance, CSL~\cite{yang2022arbitrary} reformulates the angle regression task into a discrete angle classification task, leading to various improved methods~\cite{yang2021dense,wang2022gaussian}. However, this discrete approach introduces potential calculation errors. Later methods treat the detection bounding box as a Gaussian distribution, proposing GWD~\cite{yang2021rethinking}, KLD~\cite{yang2021learning}, and KFIoU~\cite{yang2022kfiou} based on Gaussian joint optimization techniques. While these methods enhance model performance, they do not fully resolve the problem~\cite{xu2024rethinking}. Recent approaches, such as PSC~\cite{yu2023phase} and ACM~\cite{xu2024rethinking}, adopt angle encoding strategies to tackle this issue. Nonetheless, the relationship between these two methods remains unexplored, and the underlying constraint conditions are often overlooked.

\section{Restricted State Angle Resolver}
\label{sec:resolver}
\subsection{Unified Perspective of Angle Resolver}
\label{sec:theory_analyse}

Inspired by the periodic ambiguity observed in absolute phase acquisition during optical measurements, PSC~\cite{yu2023phase} introduces a phase-shifting coder technique to address the angle boundary discontinuity problem. The angle encoding states can be expressed as follows:
\begin{equation}
    m_n=\cos\left(2\theta+\frac{2n\pi}{N_\text{step}} \right),
    \label{eqn:psc}    
\end{equation}
where $n=1,2,\dots,N_{\text{step}}$ and $N_{\text{step}}$ represents the total number of encodings. Here, $\theta$ is the angle value in radians, and the range in the $le_{90}$ notation is given by $[-{\pi}/{2},{\pi}/{2})$. The model predicts all encoding states $M=\{m_1,\dots,m_n\}$ independently and decodes them accordingly~\cite{yu2023phase}.

Inspired by PSC, ACM~\cite{xu2024rethinking} introduces a coding function based on the complex exponential function. The angle encoding states can be expressed as follows:
\begin{equation}
    m=e^{j\omega\theta}=\cos(\omega\theta)+j\sin(\omega\theta),
    \label{eqn:acm}    
\end{equation}
where $m$ denotes encoded value, $j$ is the imaginary unit, $\theta$ is the angle value, and $\omega\in \mathbb{R}^+$ is the angular frequency.

The two methods utilize different forms of angle encoding, with ACM explicitly stating that these forms cannot be equivalent in any scenario. Nevertheless, both the issue of angle boundary discontinuity and the two methods can be viewed from a unified and insightful perspective of \textbf{\emph{dimensional mapping}}. Table~\ref{table:dimension_mapping} shows the comparison between our unified perspective and the previous approach, which we will analyze in detail.

\textbf{One-dimensional mapping.} The core issue with angle boundary discontinuity arises from the challenge of making one-dimensional angle values equal at both ends of a defined range. To clarify, we assume the angle range is defined as $\theta \in[-{\pi}/{2}, {\pi}/{2})$ and $\varepsilon$ represents a small positive value. As illustrated in Fig.~\ref{fig:angle_resolver}, a rotated bounding box with an angle of  $\theta_1=-\pi/2+\varepsilon$ and another with an angle of $\theta_2=\pi/2-\varepsilon$ nearly overlap in their shape. However, they exhibit a significant angle distance (\ie, $|\theta_1-\theta_2|$) in one-dimensional linear space. 
This illustrates a limitation in one-dimensional space, where two angle values that are very far apart can represent rotated bounding boxes that are very close together.

\textbf{Two-dimensional mapping.} As illustrated in the left part of Fig.~\ref{fig:angle_resolver}(b), the one-dimensional angle values can be mapped onto a circle in two-dimensional space, ensuring that the values at both ends of the angle range are equal. For simplicity, we assume the cycle is a unit circle with a radius of 1. Each point on the unit circle corresponds to a value within the defined one-dimensional angle range. This encoding mapping relationship can be represented by the following parametric equations:
\begin{equation}
    m_1=\cos(\theta), \ m_2=\sin(\theta).
    \label{eqn:2-dim}    
\end{equation}

This is consistent with the form given in Eqn.~\eqref{eqn:acm} used in ACM.  Mapping one-dimensional values to a circle in two-dimensional space already effectively resolves the issue of angle boundary discontinuity. However, this mapping relationship can be further extended into higher-dimensional space such as three-dimensional space.

\textbf{Three-dimensional mapping.} There are various ways to map one-dimensional value to a circle in three-dimensional space.
For simplicity, we define the mapping range within the unit space (\ie, each dimension has a value range of $[-1, 1]$) and assume that the encoded values in each dimension follow the same distribution. The resulting constraint equations are as follows (see Appendix for detailed derivation):
\begin{equation}
    \begin{cases}
        \sum_{i=1}^{3}m_i^2=\frac{3}{2}\\
        \sum_{i=1}^{3}m_i=0\\
    \end{cases},
    \label{eqn:condition}    
\end{equation}
where $m_i$ represents the angle encoding in $i_{th}$ dimension.

A valid solution to Eqn.~\eqref{eqn:condition} can be represented by Eqn.~\eqref{eqn:psc}, as used in the PSC method. The right part of Fig.~\ref{fig:angle_resolver}(b) provides a conceptual illustration of this case.

Although it is possible to map one-dimensional values onto a circular curve in higher-dimensional space, the constraints become progressively more complex as the number of dimensions increases. To simplify the calculations, we limit our focus to two-dimensional mapping and three-dimensional mapping.

\begin{figure*}[t]
    \centering
    \includegraphics[width=1.0\linewidth]{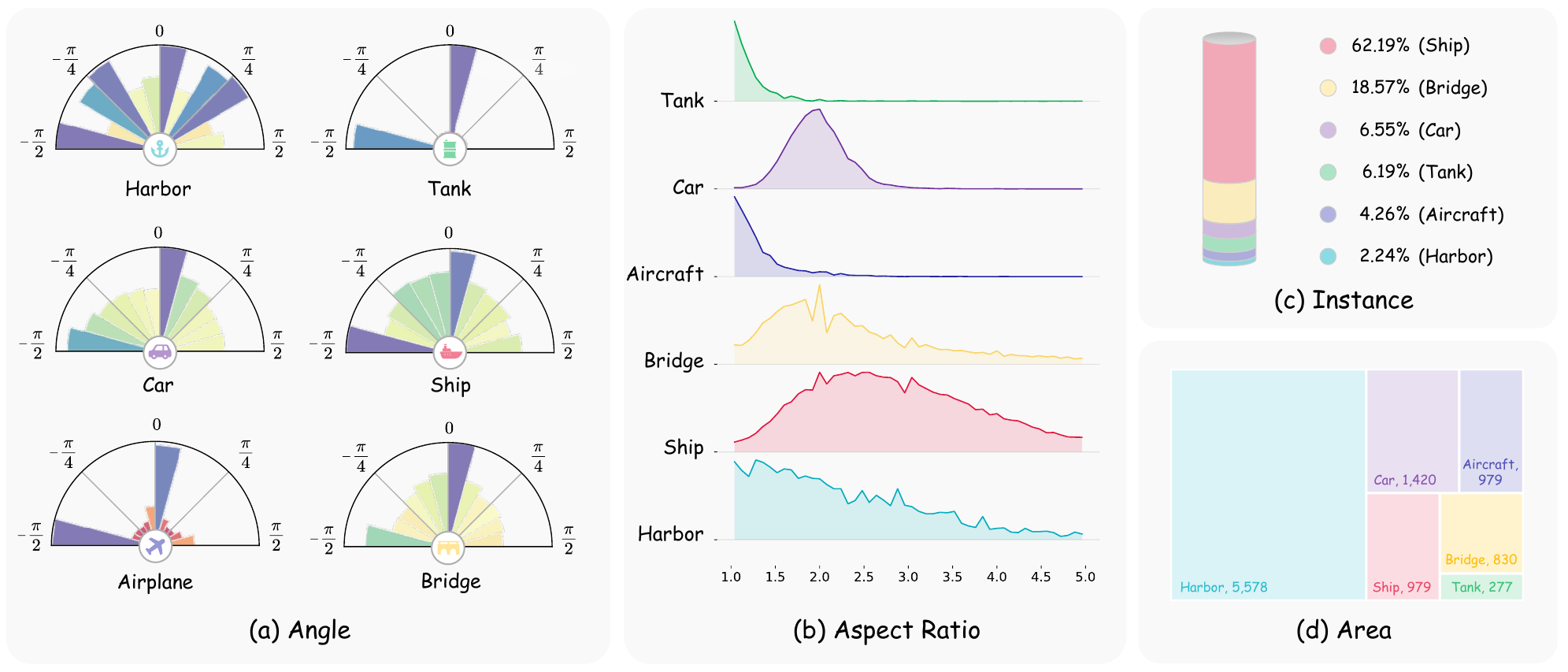}
    \caption{Statistical visualization of attributes for each category in RSAR. (a) The angle distribution of instances for each category (expressed in $le_{90}$ angle notation). (b) The aspect ratio distribution of instances for each category. (c) The percentage of instances for each category. (d) The average instance pixel area for each category.}
    \label{fig:dataset_statistics}
\end{figure*}

\begin{table}[t]
    \centering
    \setlength{\tabcolsep}{3.0mm}
    \resizebox{1\linewidth}{!}
    {
        \begin{tabular}{ll}
        \toprule
        Unified Perspective (\textbf{ours})  & Independent Perspective\\
        \hline
        one-dimension mapping & - \\
        two-dimension mapping & ACM~\cite{xu2024rethinking} (complex exponential)  \\
        three-dimension mapping & PSC~\cite{yu2023phase} (absolute phase) \\
        
        \bottomrule
        \end{tabular}
    }
    \caption{A comparison between our unified perspective and the previous method's independent perspective. \rewrite{Existing approaches (\eg, ACM~\cite{xu2024rethinking}, PSC~\cite{yu2023phase}) could be viewed as a case of our unified perspective.} }
    \label{table:dimension_mapping}
\end{table}

\subsection{Restricted State Angle Resolver}
Through this analysis, we provide a clear explanation of the angle boundary discontinuity problem from the perspective of \emph{dimensional mapping}, while also clarifying the mathematical principles behind existing methods.
From this unified perspective, we reveal the limitation of previous approaches, where each angle encoding state is predicted independently (as shown in Fig.~\ref{fig:angle_resolver}(c)). These predicted encoding states may not adhere to the inherent constraints of the unit circle. Without this constraint, a many-to-one mapping relationship exists between angle encoding states and angle values. Specifically, two angle encoding states that follow a linear scaling relationship could correspond to the same angle. This unrestricted solution space complicates the optimization of the model. 

Building on the analysis above, we introduce a new angle resolver, termed the \textbf{U}nit \textbf{C}ircle \textbf{R}esolver (\textbf{UCR}). This approach facilitates \emph{dimensional mapping} by satisfying the necessary constraints. We introduce the unit circle constraint loss to restrict the angle encoding state space of the network's predictions. To simplify the calculations, we focus on two-dimensional mapping and three-dimensional mapping.
We denote $n$ as the dimension of mapping, and the form of this loss can be expressed as:
\begin{equation}
    \mathcal{L}_{uc}=\left| \frac{n}{2}-\sum_{i=1}^{n}m_i^2\right|+\sigma(n)\left|\sum_{i=1}^n m_i\right|,
    \label{eqn:uc_loss}    
\end{equation}
where $\sigma(n)$ is a piecewise function can be expressed as:
\begin{equation}
    \sigma(n)=\begin{cases}
        1,& n=3\\
        0,& n=2
    \end{cases}.
    \label{eqn:sigma}
\end{equation}

\append{Since the model's initial predictions for the angle encoding are relatively random, and the encoding values near the center of the unit circle exhibit significant fluctuations in the early stages, we introduce an invalid region to enhance training efficiency. Within this region, the angle encoding prediction is subject only to the constraint of Eqn.~\eqref{eqn:uc_loss}, without being influenced by the angle regression constraint. This region is defined as:}
\begin{equation}
    \sum_{i=1}^nm_i^2<m_{invalid}.
\end{equation}

For the rotated object detection tasks, the total loss can be written as:
\begin{equation}
    \mathcal{L}=\mathcal{L}_{cls}+\lambda_{reg}\mathcal{L}_{reg}+\lambda_{uc}\mathcal{L}_{uc},
    \label{eqn:total_loss}
\end{equation}
where $\mathcal{L}_{cls}$ is the loss of classification and  $\mathcal{L}_{reg}$ is the loss of bounding box positioning regression. $\lambda_{reg}$ and $\lambda_{uc}$ denote the loss weight coefficient of the corresponding loss.

Our UCR can be applied for critical angle prediction in the current weakly supervised model, H2RBox-v2~\cite{yu2024h2rbox}. We leverage this model to generate rotated pseudo-labels, which assist in annotating a rotated SAR object dataset.

\section{RSAR Dataset}
\label{sec:dataset}
SARDet-100K~\cite{li2024sardet} consolidates 10 typical SAR datasets for standardization and unification. Building on this, we construct the rotated SAR object detection dataset, RSAR.

\subsection{Dataset Annotation}
\noindent{\textbf{Data Cleaning:}} Although SARDet-100K has processed the data, duplicate and unannotated images remain. Therefore, we first clean the original dataset before annotation. Specifically, we retain only one image with the maximum number of annotated objects in the case of repeated images and remove any images that lack annotations. This data-cleaning process helps prevent evaluation bias caused by dataset leaks.

\noindent{\textbf{Data Annotation:}}
We use the horizontal bounding boxes from the dataset to train the weakly supervised model~\cite{yu2024h2rbox} using our UCR. This model is then employed to generate rotated bounding boxes as references for successive manual calibration. To ensure fairness in subsequent experiments, both the test and validation sets are fully manually annotated \implus{without reference from the weakly supervised model}. It is worth noting that the orientation information for certain aircraft parts in the dataset was too ambiguous for accurate annotation, so we excluded this portion of the data.

\noindent{\textbf{Unified Format:}} We convert the annotations to DOTA format, which is easily compatible with mainstream detection frameworks.

\begin{table}[t]
    \centering
    \resizebox{1\linewidth}{!}
    {
        \begin{tabular}{ccrrcc}
        \toprule
        Dataset &  Categories & Images & Instances  & RBB \\
        \hline
        HRSID~\cite{wei2020hrsid} & 1~~ & 5,604 & 16,969  & \\
        SAR-Ship~\cite{wang2019sar} & 1~~ & 39,729 & 50,885 & \\
        SAR-AIRcraft~\cite{zhirui2023sar} & 1$^{*}$ & 4,368 & 16,463  & \\
        SIVED~\cite{lin2023sived} & 1~~ & 1,044 & 12,013  & \\
        SARDet-100K~\cite{li2024sardet} & 6~~ & 116,598 & 245,653  & \\
        RSDD-SAR~\cite{congan2022rsdd} & 1~~ & 7,000 & 10,263 &  $\checkmark$\\
        SSDD~\cite{zhang2021sar} & 1~~ & 1,160 & 2,587  & $\checkmark$\\
        SRSDD~\cite{lei2021srsdd} & 1$^{*}$ & 666 & 2,884  & $\checkmark$\\
        OGSOD~\cite{wang2023category} & 3~~ & 18,331 & 48,589  & $\checkmark$\\
        DSSDD~\cite{hu2021dual} & 1~~ & 1,236 & 3,540 & $\checkmark$\\
        \rowcolor{tablecolor} \textbf{RSAR~(ours)} & 6~~ & 95,842 & 183,534  & $\checkmark$\\
        
        \bottomrule
        \end{tabular}
    }
    \caption{Comparison of existing SAR object detection datasets with RSAR. RBB indicates rotated bounding boxes are available, and $*$ denotes categories containing fine-grained subcategories. RSAR is the largest multi-class large-scale rotated SAR object detection dataset so far.}
    \label{table:datasets_compare}
\end{table}

\begin{table*}[t]
    \centering
    \resizebox{1\linewidth}{!}
    {
        \begin{tabular}{cccc|ccc|ccc|ccc}
        \hline
        \multirow{2}{*}{Method} & \multirow{2}{*}{Resolver} & \multirow{2}{*}{DM}  & \multirow{2}{*}{Supervised}& \multicolumn{3}{c|}{RSAR } & \multicolumn{3}{c|}{DOTA-v1.0~\cite{xia2018dota} } & \multicolumn{3}{c}{HRSC~\cite{liu2017high}}\\ \cline{5-7} \cline{8-10} \cline{11-13}
         &   & & & AP$_{50}$ & AP$_{75}$ & mAP  & AP$_{50}$ & AP$_{75}$ & mAP & AP$_{50}$ & AP$_{75}$ & mAP \\
        
        \Xhline{1pt}
        FCOS (R-50)~\cite{tian2022fully} & - & - & RBB & 66.66 & 31.45 & 34.22 & 71.44 & 41.76 & 41.81 & 89.26 & 77.47 & 63.21\\ 
        \hline
        H2RBox~\cite{yang2022h2rbox} & - & - & HBB & 49.92 & 11.09 & 18.29 & 67.31 & 32.78 & 35.92 & ~~7.90 & ~~0.37 & ~~1.72\\
        H2RBox-v2~\cite{yu2024h2rbox} & ACM~\cite{xu2024rethinking} & \textcolor{dm_green}{\textbf{2}} & HBB & 65.34 & 23.53 & 30.64 & 72.37 & 40.17 & 41.05 & 89.58 & 72.02 & 58.95\\
        H2RBox-v2~\cite{yu2024h2rbox} & PSC~\cite{yu2023phase} & \textcolor{dm_purple}{\textbf{3}} & HBB & 65.16 & 24.07 & 30.91 & 72.31 & 39.49 & 40.69 & 89.30 & 64.80 & 57.98\\
        \rowcolor{tablecolor} H2RBox-v2~\cite{yu2024h2rbox} & \textbf{UCR} & \textcolor{dm_green}{\textbf{2}} & HBB & \textbf{69.21} & \underline{24.68} & \underline{32.25} &\underline{73.22} & \underline{42.26} & \underline{42.65} & \underline{89.73} & \underline{74.80} & \underline{60.00}\\
        \rowcolor{tablecolor} H2RBox-v2~\cite{yu2024h2rbox} & \textbf{UCR} & \textcolor{dm_purple}{\textbf{3}} & HBB & \underline{68.33} & \textbf{26.17} & \textbf{32.64} & \textbf{73.99} & \textbf{42.10} & \textbf{43.10} & \textbf{89.74} & \textbf{75.49} & \textbf{61.74}\\
        
        \hline
        \end{tabular}
    }
    \caption{The performance of our method and previous approaches on weakly supervised tasks across multiple datasets. DM refers to the dimension of mapping, RBB denotes the rotated bounding box, and HBB denotes the horizontal bounding box. Our method demonstrates significant improvements across all metrics compared to prior methods, which further validate the effectiveness of our unified analysis and the proposed UCR. The best score is in \textbf{bold} and the second-best is in \underline{underline}.
    }
    \label{table:weakly}
\end{table*}

\subsection{Dataset Analysis}

Our RSAR dataset comprises 95,842 images, 183,534 instances, and 6 typical object categories: \textbf{Ship, Tank, Bridge, Aircraft, Harbor, and Car}. The visualization of randomly sampled images from the dataset is shown in Fig.~\ref{fig:dataset_visual}. It is evident that, compared to horizontal bounding box annotations, rotated bounding box annotations offer greater accuracy, while also introducing more challenges for the rotated SAR object detection task. To the best of our knowledge, RSAR is the first large-scale, multi-class rotated SAR object detection dataset.

To support the advancement of rotated SAR object detection, we analyze the annotation attributes for each category in the RSAR dataset.
\textbf{(1) Angle:} We converted the angle values to the $le_{90}$ notation ($\theta \in [-\pi/2,\pi/2)$) and visualize the angle distribution for each category in Fig.~\ref{fig:dataset_statistics}(a). Notably, tanks are circular objects, so only horizontal annotations exist, consistent with the settings of previous datasets~\cite{xia2018dota}.
\textbf{(2) Aspect Ratio:} The aspect ratio of the objects also impacts detection accuracy. An aspect ratio close to 1 can introduce the square-like problem~\cite{yang2021rethinking}, while objects with an excessively high aspect ratio increase detection difficulty. Fig.~\ref{fig:dataset_statistics}(b) illustrates the distribution of the aspect ratio for each category.
\textbf{(3) Instances:} Fig.~\ref{fig:dataset_statistics}(c) displays the proportion of instances for each category, with ships accounting for the largest share. It is evident that previous studies have predominantly focused on detecting ships.
\textbf{(4) Area:}  Fig.~\ref{fig:dataset_statistics}(d) presents the average pixel area size for each category.

\subsection{Comparison with Existing Datasets}
Table~\ref{table:datasets_compare} presents the comparison between RSAR and other existing SAR object detection datasets. Previous datasets typically focus on a single object category, such as Ship (\eg, HRSID, SAR-Ship) or Aircraft (\eg, SAR-AIRcraft). However, single-category and single-scene detection data can introduce bias into specific models, undermining the evaluation of model generalization. In contrast, RSAR encompasses 6 categories of detection objects, making it the largest multi-category dataset among existing SAR object detection datasets. While the data cleaning process led to a slight reduction in the number of images and instances compared to SARDet-100K, RSAR remains a COCO-level object dataset in terms of scale.
In summary, RSAR is the first large-scale multi-class rotated SAR object detection dataset, setting it apart from existing datasets.

\section{Experiments}
\label{sec:experiments}
\noindent{\textbf{Optical Datasets:}}
\textbf{DOTA-v1.0}~\cite{xia2018dota} is a large optical aerial object detection dataset comprising 2,806 images, 15 object categories, and 188,282 instances. Following common practice~\cite{Li_2024_IJCV,yu2024h2rbox}, we split the image into 1,024 $\times$ 1,024 patches with an overlap of 200 pixels. All algorithms are trained on the training and validation sets, with evaluation conducted on the online test set.
\textbf{HRSC}~\cite{liu2017high} is an optical remote sensing ship detection dataset consisting of 1,061 images containing 2,976 instances of ships. We utilize the pre-processing provided by MMRotate~\cite{zhou2022mmrotate}, where the images are scaled to 800 × 800 for both training and testing. 

\begin{table}[t]
    \centering
    \setlength{\tabcolsep}{4.0mm}
    
    \resizebox{1.0\linewidth}{!}
    {
        \begin{tabular}{l|ccc}
        \hline
         Method &  AP$_{50}$ & AP$_{75}$ & mAP  \\
        \hline
        \rowcolor{tablecolor}\multicolumn{4}{l}{$\blacktriangledown$ \textit{One-stage}}\\
        \hline
        RetinaNet~\cite{tian2022fully} & 57.67 & 22.72 & 27.65\\
        R$^3$Det~\cite{yang2021r3det} & 63.94 & 25.02 & 30.50\\
        S$^2$ANet~\cite{han2021align} & 66.47 & 28.52 & 33.11\\
        FCOS~\cite{lin2017focal} & 66.66 & 31.45 & 34.22\\
        
        \hline
        \rowcolor{tablecolor}\multicolumn{4}{l}{$\blacktriangledown$ \textit{Two-stage}}\\
        \hline
        Faster RCNN~\cite{ren2016faster} & 63.18 & 24.88 & 30.46\\
        O-RCNN~\cite{xie2021oriented} & 64.82 & 32.69 & 33.62\\
        ReDet~\cite{han2021redet} & 64.71 & 32.84 & 34.30\\
        RoI-Transformer~\cite{ding2019learning} & 66.95 & 32.65 & 35.02\\
        \hline
        \rowcolor{tablecolor}\multicolumn{4}{l}{$\blacktriangledown$ \textit{DETR-based}}\\
        \hline
        Deformable DETR~\cite{zhudeformable} & 46.62 & 13.06 & 19.63\\
        ARS-DETR~\cite{zeng2024ars} & 61.14 & 28.97 & 31.56\\
        \hline
        \end{tabular}
    }
    \caption{The fully supervised performance in various detectors on RSAR. All models use the default configuration in MMRotate.}
    \label{table:detector}
\end{table}

\noindent{\textbf{Evaluation Metric:}}
We use average precision (AP) as the evaluation metric widely adopted in this field. Following previous methods~\cite{zeng2024ars}, we employ AP$_{75}$ as the primary evaluation metric to more effectively evaluate the accuracy of angle prediction in rotated object detection. Additionally, we include mAP and AP$_{50}$ as secondary evaluation metrics.

\noindent{\textbf{Training Details:}}
To ensure a fair comparison, all methods are implemented using the MMRotate~\cite{zhou2022mmrotate} framework, which is built on PyTorch~\cite{paszke2019pytorch}. All experiments are conducted on NVIDIA RTX 3090 GPUs, with 4 GPUs used for training on the RSAR dataset and 1 GPU for the other datasets. The DOTA-v1.0 and RSAR datasets are trained with a batch size of 2 for 12 epochs, while the HRSC dataset is trained with a batch size of 2 for 72 epochs.
For weakly supervised tasks, we utilize horizontal bounding boxes as the supervision signal, while allowing the model to predict rotated bounding boxes. We adopt the previous state-of-the-art method, H2RBox-v2~\cite{yu2024h2rbox}, as the baseline model, which employs the FCOS~\cite{tian2022fully} detector with a ResNet50~\cite{he2016deep} backbone and an FPN~\cite{lin2017feature} neck. 
All models are trained using AdamW optimizer~\cite{loshchilov2017decoupled}, with an initial learning rate of 1e-5. 

\subsection{Unit Cycle Resovler}
Building on the previous HBox-supervised state-of-the-art model, H2RBox-v2, we evaluate the performance of our resolver against earlier methods across multiple datasets, including RSAR, DOTA-v1.0, and HRSC. The results, presented in Table~\ref{table:weakly}, show significant improvements across all evaluation metrics compared to prior approaches. On the small-scale HRSC dataset, our method achieves the most pronounced improvement, with increases of 3.47\% in AP$_{75}$ and 2.79\% in mAP metrics. On the large-scale DOTA-v1.0 optical dataset, our approach enhances the AP$_{75}$ metric by 1.93\% over previous methods. Remarkably, for the first time on the DOTA-v1.0 dataset, our method achieves weakly supervised performance that exceeds that of fully supervised models using rotated bounding boxes. This demonstrates the effectiveness of our approach and establishes a strong baseline for rotated SAR object detection. On our RSAR dataset, we also observe an improvement of \append{2.10\%} in the AP$_{75}$ metric. Despite the challenges posed by complex environmental information associated with SAR images, our method effectively reduces the performance gap between weakly and fully supervised scenarios. These results further demonstrate that the weak supervision model in our method can effectively support dataset annotation.

\begin{table}[t]
    \centering
    \resizebox{1\linewidth}{!}
    {
        \begin{tabular}{c|ccc|ccc}
        \hline
         \multirow{2}{*}{$\lambda_{uc}$}& \multicolumn{3}{c|}{RSAR} & \multicolumn{3}{c}{DOTA-v1.0~\cite{xia2018dota}}\\ \cline{2-4} \cline{5-7}
         &  AP$_{50}$ & AP$_{75}$ & mAP &  AP$_{50}$ & AP$_{75}$ & mAP \\
        \hline
        0    & 65.16 & 24.07 & 30.91 & 72.31 & 39.49 & 40.69\\
        0.01 & 68.01 & 24.10 & 32.40 & 73.35 & 41.25 & 41.99\\
        0.03 & \textbf{68.45} & 23.67 & 32.03 & \textbf{73.99} & \textbf{42.10} & \textbf{43.10}\\
        0.05 & 68.33 & \textbf{26.17} & \textbf{32.64} & 73.81 & 41.99 & 42.38\\
        0.1  & 67.62 & 25.86 & 32.58 & 73.17 & 42.03 & 42.13\\
        0.2  & 66.80 & 25.10 & 31.94 & 73.20 & 41.75 & 41.72\\
        \hline
        \end{tabular}
    }
    \caption{Ablation experiments on different loss weights for unit cycle loss. The best performance on RSAR is achieved with a loss weight of 0.05, and the optimal loss weight for DOTA-v1.0 is 0.03.}
    \label{table:ablation_hyperparameter}
    \vspace{-5pt}
\end{table}

\begin{table}[t]
    \centering
    \setlength{\tabcolsep}{3.0mm}
    
    \resizebox{1.0\linewidth}{!}
    {
        \begin{tabular}{cc|ccc}
        \hline
        unit cycle loss & invalid region & AP$_{50}$ & AP$_{75}$ & mAP\\
        \hline
        & & 72.37 & 40.17 & 41.05\\
        & \checkmark & 72.68 & 41.57 & 42.09\\
        \checkmark & &73.16 & 40.94 & 41.68\\
        \checkmark & \checkmark & \textbf{73.22} & \textbf{42.26} & \textbf{42.65}\\
        
        \hline
        \end{tabular}
    }
    \caption{\append{Ablation experiments on various strategies in the two-dimensional mapping of UCR. Optimal results are achieved by combining the unit circle loss with the invalid areas.}}
    
    \label{table:ablation_region}
    \vspace{-10pt}
\end{table}

\subsection{Performance of RSAR in Various Detector}
To better demonstrate the performance and characteristics of our RSAR dataset, we conduct experiments using mainstream one-stage, two-stage, and DETR-based rotated object detection frameworks. The experimental results are presented in Table~\ref{table:detector}. FCOS achieves the best performance among the one-stage frameworks, while RoI-Transformer outperforms others in the two-stage category. Overall, the two-stage frameworks deliver higher accuracy than other frameworks. Additionally, RSAR exhibits lower performance across various detectors when compared to the optical dataset DOTA-v1.0, highlighting the challenges of the rotated SAR object detection task.

\begin{figure*}[t]
    \vspace{-4pt}
    \centering
    \includegraphics[width=0.98\linewidth]{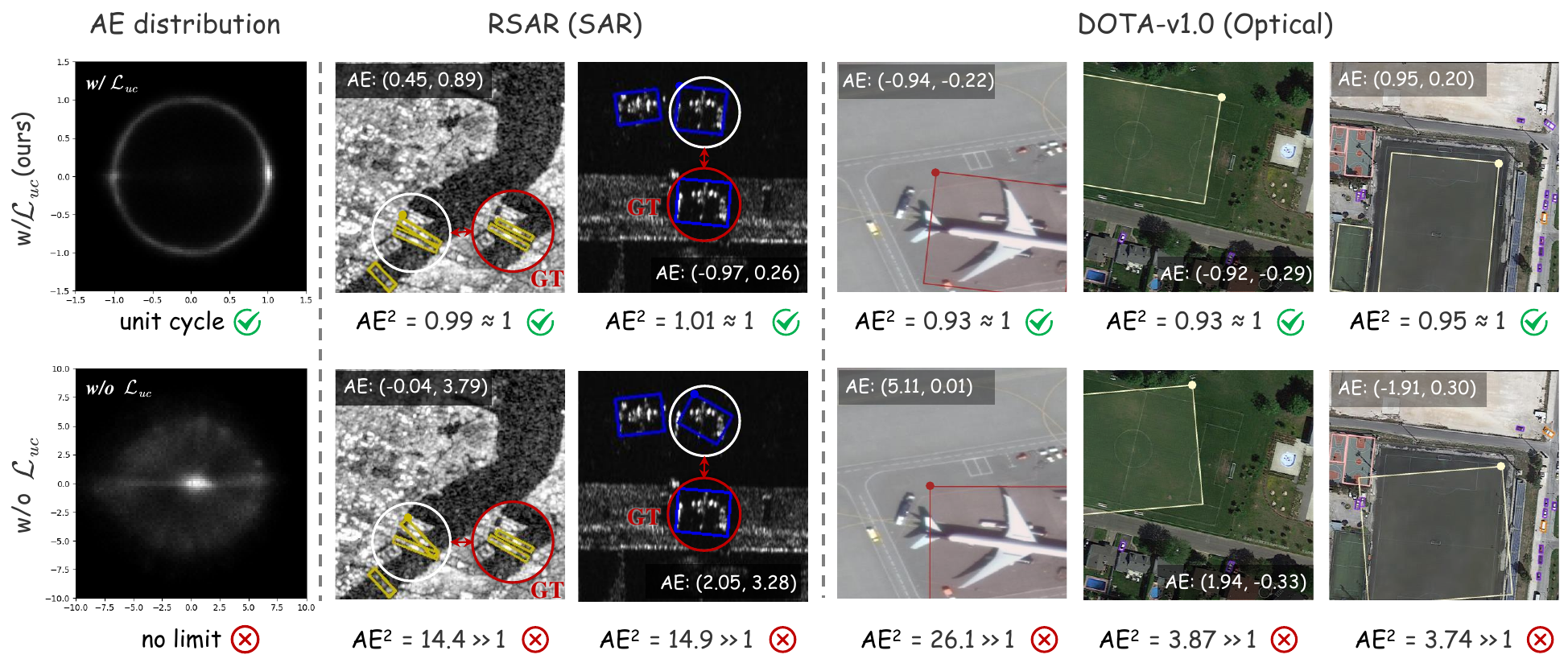}
    \vspace{-6pt}
    \caption{Comparison of visualized results on RSAR and DOTA-v1.0 in two-dimensional mapping. AE represents angle encoding, indicating the $(\cos\theta, \ \sin\theta)$ of the model's prediction, while AE$^2$ denotes their sum of squares (\ie, $AE^2=\sin^2\theta+\cos^2\theta$). We obtain the predicted values for all angle encodings of the bounding boxes on the test set and display their probability distribution statistics in the image on the left. The white regions in the probability distribution diagram correspond to areas where angle encoding has a higher probability of occurrence. Due to the unrestricted distribution in the angle encoding state space of the no-limit method, angle predictions may lack accuracy in certain scenarios. Our method significantly enhances the accuracy of angle prediction in the model.}

    \label{fig:vis_result}
    \vspace{-10pt}
\end{figure*}

\subsection{Ablation Studies}
\noindent{\textbf{Loss weights:}}
Table~\ref{table:ablation_hyperparameter} illustrates the impact of different loss weights for the unit cycle loss on model performance in weakly supervised tasks. The model achieves optimal performance on the RSAR dataset with a weight of 0.05, while the best results on the DOTA-v1.0 dataset are obtained with a weight of 0.03. Notably, compared to the baseline where our method is not applied ($\lambda_{uc}=0$), our approach significantly enhances both AP$_{75}$ and mAP metrics, even with a small loss weight. As the loss weight increases, the model's performance begins to decline. This may be due to the fact that overly stringent constraints hinder the model's ability to learn angles effectively. Overall, these results highlight the effectiveness and robustness of our method.

\noindent{\textbf{Invalid region:}}
\append{In our approach, we employ the unit circle loss to constrain the representation space of angle encoding and define the region near the center of the unit circle as an invalid region. Within this region, angle encoding is supervised solely by the unit circle loss and is exempt from angle regression supervision. Table~\ref{table:ablation_region} highlights the impact of these two strategies on model performance. The results show that simply defining the invalid region significantly improves performance, indicating that angle prediction in this region negatively affects the model and underscoring the importance of restricting the angle space. Overall, both strategies enhance model performance, with their combined use yielding even greater benefits.}

\begin{table}[t]
    \centering
    \setlength{\tabcolsep}{2.3mm}
    
    \resizebox{1.0\linewidth}{!}
    {
        \begin{tabular}{c|ccc|ccc}
        \hline
         \multirow{2}{*}{Loss} & \multicolumn{3}{c|}{RSAR} & \multicolumn{3}{c}{DOTA-v1.0~\cite{xia2018dota}}\\ \cline{2-4} \cline{5-7}
         & AP$_{50}$ & AP$_{75}$ & mAP &  AP$_{50}$ & AP$_{75}$ & mAP \\
        \hline
        L1& 68.33 & 26.17 & 32.64 & 73.99 & 42.10 & 43.10\\
        MSE& 68.56 & 24.27 & 32.03 & 73.89 & 41.89 & 42.67\\
        \hline
        \end{tabular}
    }
    \caption{Ablation experiments on different loss functions for unit cycle loss. Higher performance gains can be achieved using L1 loss functions compared to MSE loss functions.}
    \label{table:ablation_loss_function}
    \vspace{-5pt}
\end{table}

\noindent{\textbf{Loss Function:}}
In our method, we calculate the unit cycle loss between the model predictions and the inherent constraints. To assess the impact of different loss functions on the model performance, we conducted an ablation experiment, the results of which are presented in Table~\ref{table:ablation_loss_function}.
The use of L1 loss functions yields higher performance gains compared to MSE loss functions. The similar experimental performance for different loss functions also demonstrates the stability of our method.

\noindent{\textbf{Visualization Result:}}
To better demonstrate how our method improves the model's angle predictions, we compared the visualization results of our approach with the baseline method, as shown in Fig.~\ref{fig:vis_result}. On the left, we present the distribution statistics of the model's predicted angle encoding states on the test set, where white areas indicate regions with a high probability of occurrence for these encoding states. The baseline method, which lacks constraints, shows a randomly scattered distribution. In contrast, our method exhibits a well-defined unit circle distribution, indicating that our angle predictions are more regular. In the visualizations for both the RSAR and DOTA-v1.0 datasets, we observe that the baseline model tends to produce extreme predictions for the angle encoding states (\ie, $AE^2\gg1$), which negatively impacts the final prediction results. Our method provides more accurate predictions for the angles of detection objects, as it adheres to the necessary constraints (i.e., $AE^2 \approx 1$).

\section{Conclusion}
\label{sec:conclusion}
In this paper, we address the issue of angle boundary discontinuity in rotated object detection from a unified perspective based on dimension mapping. Our analysis reveals that existing methods overlook the inherent unit cycle constraints in angle encoding, which leads to bias in angle predictions. To overcome this, we propose a unit circle resolver (UCR) that ensures angle encoding satisfies these constraint conditions, thereby improving the accuracy of angle prediction. We apply UCR in weakly supervised models to generate pseudo-rotated labels, and after manual calibration, we introduce RSAR—a large-scale, multi-class, rotated SAR object detection dataset. To our best knowledge, RSAR is the largest dataset in the field to date. Experimental results on both the RSAR and optical datasets demonstrate that our method significantly enhances angle prediction accuracy, even surpassing fully supervised models on DOTA-v1.0.

\noindent{\textbf{Limitations and future work:}}
This paper primarily focuses on the improvement of weakly supervised models and the construction of a rotated SAR object detection dataset. However, designing a fully supervised detection model tailored for SAR detection using the RSAR dataset remains an important avenue for future research.

{
    \small
    \bibliographystyle{ieeenat_fullname}
    \bibliography{main}
}

\clearpage
\setcounter{page}{1}
\maketitlesupplementary
\appendix





\section{Proof of Three-dimensional Mapping}
In Sec.~\ref{sec:theory_analyse}, we introduce constraints for three-dimensional mapping and present Eqn.~\eqref{eqn:condition}. We will provide the detailed deduction and proof in this section.

We assume that the encoded values follow the same distribution across each dimension, implying that the absolute values of the weights of each dimension are equal. To simplify, we further assume that all dimensions have identical weights. Based on this assumption, the following formula can be derived:
\begin{equation}
    m_1+m_2+m_3=0.
    \label{eqn:plane}
\end{equation}

This equation describes the two-dimensional plane in which the mapping data resides. The normal vector of the plane can be expressed as $\boldsymbol{w}=[1,1,1]^T$.

One-dimensional values are mapped into a three-dimensional unit space (\ie, each dimension has a value range of $[-1,1]$) to form a circular curve. Given this condition and the plane defined in Eqn.~\eqref{eqn:plane}, we can derive that the distance from any encoded values to the origin is equal to $\frac{3}{2}$. Thus, we derive the following formula:
\begin{equation}
    m_1^2+m_2^2+m_3^2=\frac{3}{2}.
    \label{eqn:cycle}
\end{equation}

We apply a basis transformation along with the parametric equation of a circle to derive the analytic solution. In this context, we define two unit vectors that lie on the plane: $\boldsymbol{v}=[-\frac{1}{\sqrt{2}},\frac{1}{\sqrt{2}},0]^T$ and $\boldsymbol{u}=[\frac{1}{\sqrt{6}},\frac{1}{\sqrt{6}},-\frac{2}{\sqrt{6}}]^T$. These vectors form an orthonormal basis for the plane. Using this basis, we can re-express any encoded values on the plane in terms of these vectors:
\begin{equation}
    \boldsymbol{m}=r(\boldsymbol{u}\cos\theta+\boldsymbol{v}\sin\theta).
    \label{eqn:re_express}
\end{equation}

Here, $r$ represents the polar diameter. By combining Eqn.~\eqref{eqn:re_express} and Eqn.~\eqref{eqn:cycle}, we obtain $r^2=\frac{3}{2}$. Therefore, the analytic solution for the encoded values on the plane can be expressed as:
\begin{equation}
    \begin{bmatrix}m_1\\m_2\\m_3\end{bmatrix}=\begin{bmatrix}\frac{1}{2} & -\frac{\sqrt{3}}{2}\\\frac{1}{2} & \frac{\sqrt{3}}{2}\\-1 & 0\end{bmatrix} \begin{bmatrix}\cos\theta\\\sin\theta\end{bmatrix}=\begin{bmatrix}\cos(\theta+\frac{2\pi}{3})\\\cos(\theta+\frac{4\pi}{3}) \\ \cos(\theta+\frac{6\pi}{3})\end{bmatrix}.
    \label{eqn:trans_result}
\end{equation}

Eqn.~\eqref{eqn:trans_result} aligns with Eqn.~\eqref{eqn:psc} proposed in PSC, allowing PSC to be interpreted from a unified perspective of dimensional mapping.

\begin{figure}[t]
    \centering
    \includegraphics[width=1.0\linewidth]{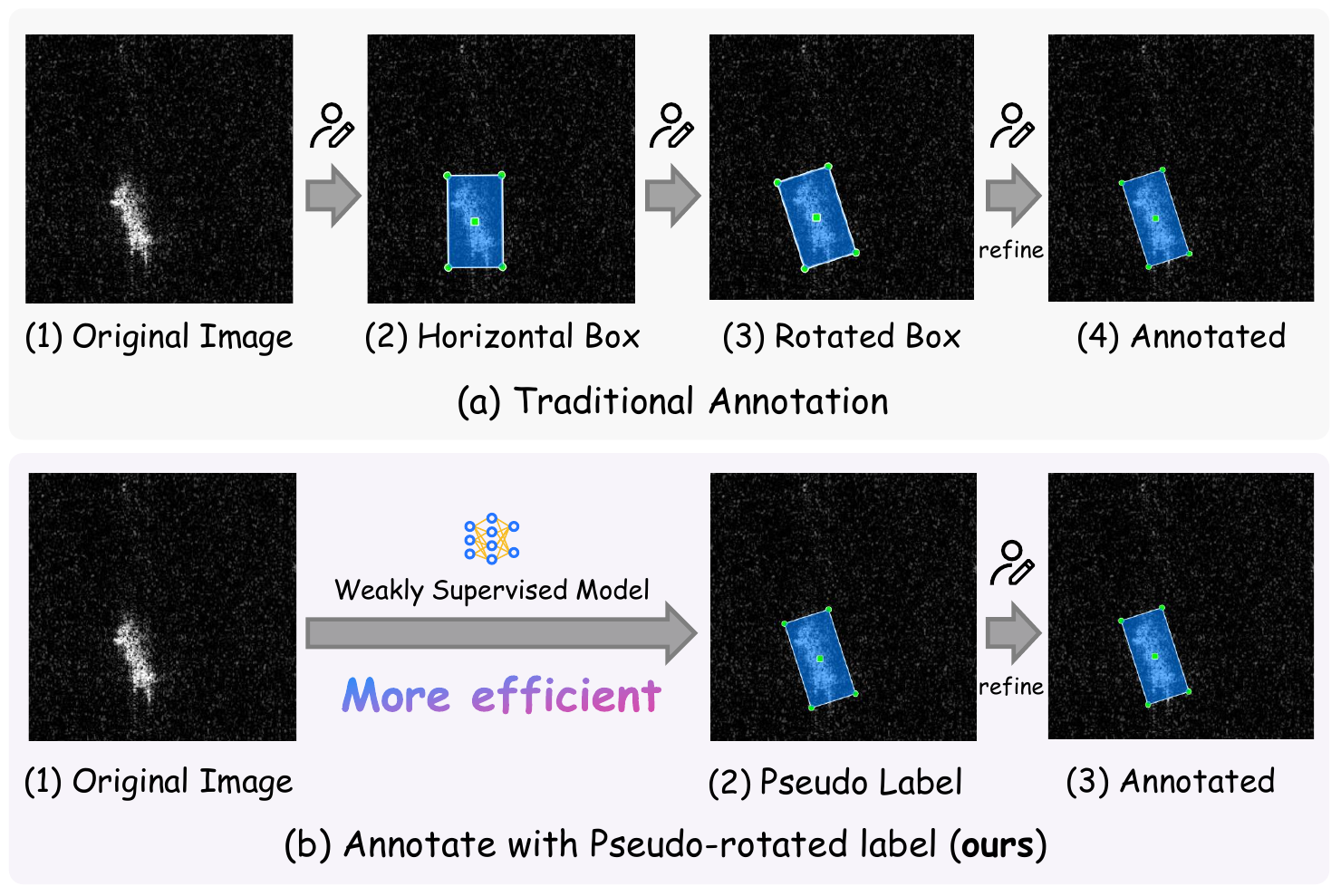}
    \caption{A comparison between the traditional annotation method and our approach, which uses pseudo-rotated labels from a weakly supervised model. It demonstrates our method simplifies and improves the efficiency of the annotation process.
    }
    \label{fig:annotation}
\end{figure}

\section{Detail in Dataset Annotation}
Fig.~\ref{fig:annotation} illustrates a comparison between our annotation method and the traditional annotation method. Notably, the general practice for rotated annotation involves using a rectangular box, requiring the adjustment to begin with a horizontal rectangle. In the traditional method, the process begins by roughly determining a horizontal bounding box based on the object's position, followed by rotating the box to align with the object's orientation and finally fine-tuning its position. In contrast, our method employs a weakly supervised model to generate pseudo-rotated labels with high accuracy, requiring only minimal manual fine-tuning, which greatly enhances annotation efficiency. 

Additionally, the angle adjustment step can be entirely skipped when the angle predictions are sufficiently accurate. Thus, this paper aims to improve the weakly supervised model's accuracy in angle prediction. To minimize the impact of pseudo-labels on manual labeling, traditional methods are used to label both the validation and test sets to avoid evaluation bias.

\begin{table}[t]
    \centering
    
    \resizebox{0.8\linewidth}{!}
    {
        \begin{tabular}{cc|ccc}
        \hline
        Resolver & DM & AP$_{50}$ & AP$_{75}$ & mAP \\
        \hline
        PSC~\cite{yu2023phase} & 4 & 71.98 & 40.03 & 41.25 \\
        \rowcolor{tablecolor} UCR (\textbf{ours}) & 4 & 73.76 & 41.62 & 42.67\\
        PSC~\cite{yu2023phase} & 5 & 72.33 & 38.69 & 39.89\\
        \rowcolor{tablecolor} UCR (\textbf{ours}) & 5 & 73.85 & 42.86 & 42.98\\
        
        \hline
        
        \end{tabular}
    }
    \caption{A comparison of previous methods with our UCR approach in higher-dimensional mappings. All experiments are based on H2RBox-v2~\cite{yu2024h2rbox}. Our method achieves superior performance in higher-dimensional mapping scenarios.}
    \label{table:higher_dimension}
\end{table}

\section{Additional Experiment Results}
\subsection{Multi-dimensional Mapping}
To simplify optimization and computation, we focus on two-dimensional mapping and three-dimensional mapping of UCR in this study. For higher-dimensional mappings, encoded values must satisfy more additional constraints. Using four-dimensional mapping and five-dimensional mapping as examples, we define the corresponding constraints and conduct experimental validation on the DOTA-v1.0 dataset.

For a four-dimensional mapping, each encoded value must satisfy the following conditions:
\begin{equation}
    \begin{cases}
\sum_{i=1}^4m_i^2=2\\
m_1+m_3=0\\
m_2+m_4=0
\end{cases}.
\end{equation}

For a five-dimensional mapping, each encoded value must satisfy the following conditions:
\begin{equation}
    \begin{cases}
\sum_{i=1}^5m_i^2=2.5\\
\sum_{i=1}^5m_i=0\\
\sum_{i=1}^5m_i^3=0\\
\end{cases}.
\end{equation}

Table~\ref{table:higher_dimension} presents the results of experiments on higher-dimensional mappings. The results indicate that our UCR achieves greater performance improvements than the previous resolvers in higher-dimensional scenarios. However, as the number of mapping dimensions increases, the constraints become more complex. Therefore, we primarily focus on two-dimensional and three-dimensional mappings.

\begin{table}[t]
    \centering
    \setlength{\tabcolsep}{2.3mm}
    
    \resizebox{1.0\linewidth}{!}
    {
        \begin{tabular}{c|ccc|ccc}
        \hline
         \multirow{2}{*}{$r^2$} & \multicolumn{3}{c|}{RSAR} & \multicolumn{3}{c}{DOTA-v1.0~\cite{xia2018dota}}\\ \cline{2-4} \cline{5-7}
         & AP$_{50}$ & AP$_{75}$ & mAP &  AP$_{50}$ & AP$_{75}$ & mAP \\
        \hline
        0.5& 67.89 & 23.96 & 31.75 & 73.18 & 41.37 & 42.05\\
        1.5& 68.33 & 26.17 & 32.64 & 73.99 & 42.10 & 43.10\\
        3.0& 68.45 & 23.37 & 32.00 & 73.78 & 41.31 & 42.03\\
        \hline
        \end{tabular}
    }
    \caption{Ablation experiment on different ranges of mapping in three-dimensional mapping. Constraining the mapping range to unit space yields better results.}
    
    \label{table:ablation_radius}
    \vspace{-5pt}
\end{table}

\subsection{The Range of Mapping}
In Sec.~\ref{sec:theory_analyse}, we mention that there are multiple ways in which one-dimensional values can be mapped to a circle in multi-dimensional space. To simplify, we restrict the mapping range to unit space (\ie, each dimension has a value range of $[-1,1]$) and present the formula for Eqn.~\eqref{eqn:condition}. If this constraint of unit space is removed, we obtain a new mapping form (\ie, $\sum_i^nm^2_i=r^2$, where $r>0$).
To validate the influence of different mapping forms on the model, we utilize three-dimensional mapping as an example and present the experimental results in Table~\ref{table:ablation_radius}. 
The experimental results indicate that the model achieves optimal performance when $r^2 = 1.5$ (\ie, unit space). A larger mapping range can lead to more dispersed encoding states, making optimization and prediction more challenging. Conversely, a smaller mapping range may cause reduced differences between encoding states, resulting in prediction biases. Overall, restricting the mapping to unit space provides a more generalized approach, resulting in better performance for angle prediction.

\begin{table}[t]
    \centering
    
    \resizebox{0.65\linewidth}{!}
    {
        \begin{tabular}{c|ccc}
        \hline
        $m_{invalid}$ & AP$_{50}$ & AP$_{75}$ & mAP \\
        \hline
        0 & 73.16 & 40.94 & 41.68 \\
        0.1 & 73.21 & 41.62 & 41.89\\
        0.2 & 73.22 & 42.26 & 42.65\\
        0.5 & 73.13 & 40.98 & 41.92\\
        1.0 & 39.56 & 7.87 & 14.42\\
        \hline
        \end{tabular}
    }
    \caption{Ablation experiments on different ranges of the invalid region conducted by two-
dimensional mapping of UCR. Optimal results are achieved when the threshold is taken as 0.2.}
    \label{table:ablation_region_range}
\end{table}

\begin{table}[t]
    \centering
    \setlength{\tabcolsep}{1.2mm}
    
    \resizebox{1.0\linewidth}{!}
    {
        \begin{tabular}{l|cccccc}
        \hline
         Method & SH & AI & CA & TA & BR & HB  \\
        \hline
        \rowcolor{tablecolor}\multicolumn{7}{l}{$\blacktriangledown$ \textit{One-stage}}\\
        \hline
        RetinaNet~\cite{tian2022fully} & 73.6 & 73.5 & 73.6 & 22.4 & 49.6 & 53.4\\
        R$^3$Det~\cite{yang2021r3det} & 78.7 & 73.2 & 89.3 & 22.6 & 56.9 & 63.0 \\
        S$^2$ANet~\cite{han2021align} & 82.3 & 77.8 & 89.8 & 25.8 & 60.2 & 63.0\\
        FCOS~\cite{lin2017focal} & 79.0 & 73.0 & 89.8 & 33.9 & 58.8 & 65.5\\
        
        \hline
        \rowcolor{tablecolor}\multicolumn{7}{l}{$\blacktriangledown$ \textit{Two-stage}}\\
        \hline
        Faster RCNN~\cite{ren2016faster} & 78.3 & 76.8 & 89.5 & 30.8 & 54.7 & 49.0\\
        O-RCNN~\cite{xie2021oriented} & 79.4 & 75.3 & 89.7 & 29.7 & 56.2 & 58.5\\
        ReDet~\cite{han2021redet} & 79.0 & 78.1 & 89.5 & 25.6 & 55.0 & 61.1 \\
        RoI-Transformer~\cite{ding2019learning} & 85.9 & 76.5 & 90.1 & 27.5  & 57.4 & 64.4\\
        
        \hline
        \rowcolor{tablecolor}\multicolumn{7}{l}{ $\blacktriangledown$ \textit{DETR-based}}\\
        \hline
        Deformable DETR~\cite{zhudeformable} & 58.0 & 51.3 & 66.5 & 21.7 & 36.8 & 45.4\\
        ARS-DETR~\cite{zeng2024ars} & 76.9 & 70.2 & 80.4 & 29.1 & 51.2 & 59.1\\
        \hline
        \end{tabular}
    }
    \caption{The detailed fully supervised performance in various detectors on RSAR. All results present AP$_{50}$ for each category.}
    \label{table:detailed_detector}
    \vspace{-10pt}
\end{table}

\subsection{The Range of Invalid Region.}
Table~\ref{table:ablation_region_range} illustrates the effect of various invalid region ranges on the performance of the weakly supervised model. The findings reveal that incorporating the invalid region enhances model learning; however, extensive invalid regions may lead to insufficient constraints for angle regression.

\subsection{Detailed Results on Fully Supervised Model}
Table~\ref{table:detailed_detector} summarizes the performance of various fully supervised models on the RSAR dataset, detailing the results for each category. The performance is evaluated using the AP$_{50}$ metric, with categories represented by their respective abbreviations: Ship (SH), Aircraft (AI), Car (CA), Tank (TA), Bridge (BR), and Harbor (HA).

\end{document}